\documentclass[letterpaper, 10 pt, conference]{ieeeconf}

\IEEEoverridecommandlockouts                              

\usepackage{amsmath} 
\usepackage{amssymb}  
\usepackage{mathrsfs} 
\usepackage{graphicx,cite}
\usepackage{enumerate}
\usepackage{comment}
\usepackage{algorithm}
\usepackage{algpseudocode}
\usepackage{tcolorbox}
\usepackage{tabularx}
\usepackage{array}
\usepackage{dsfont}
\usepackage{setspace}

\newcommand*{\QEDA}{\null\nobreak\hfill\ensuremath{\blacksquare}}

\newtheorem{rem}{Remark}
\newtheorem{prop}{Proposition}
\newtheorem{lem}{Lemma}

\newtheorem{defn}{Definition}

\newtheorem{assum}{Assumption}

\newcommand{\prob}{\ensuremath{P}}

\newcommand{\raf}[1]{{\color{red} RAF: #1}}
\newcommand{\abhijit}[1]{{\color{blue} ABHI: #1}}
\newcommand{\mnl}[1]{{\color{magenta} MNL: #1}}

\newtheorem{thm}{Theorem}

\providecommand{\keywords}[1]{\textbf{\textit{Index terms---}} #1}

\newcommand{\remend}{\relax\ifmmode\else\unskip\hfill\fi\hbox{$\bullet$}}

\title{\LARGE \bf
Safe Reinforcement Learning for Constrained Markov Decision Processes with Stochastic Stopping Time }

\author{Abhijit Mazumdar, Rafal Wisniewski and Manuela  L. Bujorianu 
\thanks{The work of the first and the second authors has been supported by the Poul Due Jensens Foundation under project SWIft.The work of the third author has been funded by the EPSRC project EP/R006865/1: Interface reasoning for interacting systems (IRIS).}
\thanks{A. Mazumdar and R. Wisniewski are with the Section of Automation $\&$ Control, Aalborg University, 9220 Aalborg East, Denmark (e-mail: \{abma, raf\}@es.aau.dk).}
\thanks{Manuela L. Bujorianu is with the Department of Computer Science, University College London, UK (e-mail: l.bujorianu@ucl.ac.uk).}}

\graphicspath{{figs/}}

\begin{document}

\maketitle
\thispagestyle{empty}
\pagestyle{empty}

\begin{abstract}
In this paper, we present an online reinforcement learning algorithm for constrained Markov decision processes with a safety constraint. Despite the necessary attention of the scientific community, considering stochastic stopping time, the problem of learning optimal policy without violating safety constraints during the learning phase is yet to be addressed. To this end, we propose an algorithm based on linear programming that does not require a process model. We show that the learned policy is safe with high confidence. We also propose a method to compute a safe baseline policy, which is central in developing algorithms that do not violate the safety constraints. Finally, we provide simulation results to show the efficacy of the proposed algorithm. Further, we demonstrate that efficient exploration can be achieved by defining a subset of the state-space called \textit{proxy set}.
 
\end{abstract}
\section{Introduction}
Reinforcement learning (RL) is a machine learning paradigm that solves sequential decision problems in a model-free setting. RL has become a popular tool for designing control policies for dynamical systems as it eliminates the need for a model of the system. Various RL algorithms have been successfully developed to solve some highly complex learning problems, such as learning how to play computer games better than human experts \cite{mnih2013playing} and making robots learn complicated tasks \cite{kober2013reinforcement}, among others. When it comes to applications of RL algorithms to safety-critical systems such as self-driving cars, power plants, etc., these algorithms must provide safe control policies even during the learning phase. One of the most challenging issues with RL is maintaining safety during the exploration phase. In principle, an online safety assessment of a control policy is only possible if the agent applies it, resulting in potential safety violations. However, with a reasonable number of visits, one can assess a safe policy with high confidence for certain safety notions, such as probabilistic safety. 
\par \textit{Related literature:}  The notion of safety that considered in this work is $p$-safety \cite{wisniewski2017stochastic,wisniewski2020p,bujorianu2020stochastic,wisniewski2021safety,wisniewski2023probabilistic}. It is defined as the probability of visiting unsafe states before visiting goal states. Among these, \cite{bujorianu2020stochastic, wisniewski2023probabilistic} have investigated $p$-safety for Markov decision processes (MDPs) and Markov chains. Recently, \cite{mazumdar2023online_ecc,mazumdar2023online} have proposed online algorithms to learn the safety function for assessing safety for MDPs. However, the last two works do not consider optimal policy design. Another notion of probabilistic safety is considered in \cite{el2018controlled} based on the probability of visiting unsafe states at a given time instant without a stopping time or terminal states. 
\par Safe RL has received massive attention from the research community working with RL. For an overview of various approaches to safe RL, one can look at \cite{garcia2015comprehensive}. The problem of safe policy design for MDPs can be cast as a constrained MDP. There are various approaches proposed for solving CMDPs, such as safety filters or shielding based approach \cite{jansen2020safe,bastani2021safe}, linear programming (LP)-based approach \cite{altman1991adaptive,jaksch2010near,efroni2020exploration,singh2022learning,bura2022dope}, Lagrangian formulation \cite{wachi2020safe,wei2022triple,wei2022provably,wei2023provably}, policy gradient-based approach \cite{yu2019convergent,ding2020natural,papini2022smoothing,guin2023policy}.
In most of these works, the safety constraints are not guaranteed to be satisfied during the learning phase. Among these works, \cite{bura2022dope,wei2022triple,wei2022provably} maintains zero constraint violation during the learning with high confidence. All the above works consider a deterministic finite or infinite horizon. 
\par \textit{Contributions:}   Our main objective is to design a decision policy for an MDP that maximizes a cumulative reward while maintaining safety throughout the learning phase. To begin with, we partition the state-space into three subsets: a taboo set, a forbidden set or unsafe set, and a target or goal set. The target set contains the goal states, and the forbidden set contains the unsafe states that need to be avoided with a given probability. Put simply, we seek to design a policy that results in reaching the target set before reaching the forbidden set with a probability less than or equal to $p$. The process is stopped whenever it reaches either the forbidden states or the goal states. We assume that the stopping time is stochastic and almost surely finite. In other words, the forbidden or unsafe states and the goal states are terminal. We can model many real-world examples with this setup. 
To illustrate, our setup captures a robot's task of reaching the goal while avoiding objects on its path to the goal. This framework is the so-called reach-avoidance problem. At the outset, we formulate the safe RL problem as a constrained MDP. It is well known that, for constrained MDP, one can not leverage dynamic programming as the principle of optimality does not hold \cite{haviv1996constrained}. However, it is possible to solve the problem by formulating it as a linear program (LP) if the transition probabilities are known \cite{wisniewski2023probabilistic}. 
We present an algorithm, namely \textit{$p$-safe RL}, based on the LP formulation, ensuring that the safety constraint is not violated with high confidence. Since we consider finite stopping time, we design an episodic algorithm. Our algorithm learns the optimal safe policy while not visiting the forbidden set with a probability more than a prescribed threshold. Finally, we showcase the efficacy of our proposed algorithm using one example MDP. We summarize the main contributions of the paper as follows.
\begin{itemize}
\item At the outset, we shed some light on the need to use a stopping time instead of a fixed horizon. For safety with a fixed horizon, an unsafe state will eventually be visited even though the safety constraint is satisfied with probability one.   
    \item We present an algorithm that considers stochastic stopping time and guarantees $p$-safety during the exploration phase. Our algorithm is based on the \textit{optimism in the face of uncertainty (OFU)} idea \cite{jaksch2010near,efroni2020exploration}. While OFU offers a better reward bonus, it does not guarantee satisfaction of the safety constraints. To this end, based on the OFU principle, we propose a safe RL algorithm.
    \item We derive a closed-form expression for a safe baseline policy. This is crucial in designing safe RL algorithms that do not violate the safety constraint during the learning phase. In the existing works, it is generally assumed that a safe baseline policy is known. 
    \item We define a subset of the state-space called \textit{proxy set} that contains all the states from where it is possible to visit the forbidden states. With simulation, we demonstrate that knowledge of these \textit{proxy states} results in learning better policy than the policy designed without knowing these states.   
\end{itemize}
\par The paper is structured in the following manner. Section \ref{safety-MDP} presents the problem and mathematical background required for the paper. In Section \ref{sec_Algo}, we describe the algorithm in detail along with all the main results. We provide simulation results to demonstrate the empirical efficacy of the proposed algorithm in Section \ref{example}. Finally, in Section \ref{sec_conclusion}, we conclude the paper by highlighting some future extensions. 
\section{Background and Problem Formulation:} \label{safety-MDP}
We consider an MDP with the state-action space $(\mathcal{X}, \mathcal{A})$, where $\mathcal{X}$ is a finite set of states and $\mathcal{A}$ is a finite set of actions. We denote by $\Omega$ the sample space of all sequences of the form $\omega = (x_0,a_0,x_1,a_1, \hdots) \in (\mathcal{X} \times \mathcal{A})^{\infty}$ where $x_i \in \mathcal{X}$ and $a_i \in \mathcal{A}$. By $\mathcal{F}$, we denote the $\sigma$-algebra generated by coordinate mappings: $X_t(\omega) = x_t$ and $A_t(\omega) = a_t$. We consider stochastic policies that depend only on the initial state. For a fixed initial state $x_0$, it is a map $\pi^{{x}_0}: \mathcal{X} \to \Delta(\mathcal{A})$, where $\Delta(\mathcal{A})$ is the probability simplex over the action space, i.e., $\Delta(\mathcal{A}) = \{(P_1, \hdots, P_{|\mathcal{A}|}) \in [0,1]^{\mathcal{A}} \mid P_1 + \hdots + P_{|\mathcal{A}|} = 1\}.$ 
By $\mathcal{D}^{x_0}=\{\pi^{x_0}: \mathcal{X} \to \Delta(\mathcal{A}) | X_0=x_0\}$, we denote the set of policies that depend on initial state. For a set $Y$, by $|.|$, we denote the cardinality of $Y$, and for a scalar $y$, we denote the absolute value by $|y|$. The transition probabilities of the MDP are time invariant and are defined as $\prob(x,a,y):= \mathds{P}[X_{t+1}=y|X_t=x,A_t=a]$ for all $x\in \mathcal{X}$, $y\in \mathcal{X}$ and $a\in \mathcal{A}$. Suppose $r: \mathcal{X}\times \mathcal{A}\rightarrow \mathds{R}$ denotes the reward associated with each transition from the state $x$ upon applying an action $a$, i.e., the reward is $r(x,a)$. The map $r$ induces a process $(r_t)$ given as below:
\begin{align*}
    r_t = r(X_t,A_t).
\end{align*}
We partition the state-space into a target or goal set $E$, a set of forbidden states $U$, and $H := \mathcal{X} \setminus (E\cup U)$ be the set of taboo state. The goal is to reach the target set $E$ before reaching the forbidden set $U$. For each state in $x\in \mathcal{X}$, the safety function corresponds to the  probability that the realizations hit the forbidden set $U$ before the target set $E$ is reached if the process starts from $x$. Following \cite{wisniewski2023probabilistic}, we define the safety function for each state $x\in H$ as follows:
\begin{align*}
    S^P_{\pi^{x_0}}(x) := \mathds{P}^{x,P}_{\pi^{x_0}}[\tau_U < \tau_{E}],
\end{align*}
 where, $\tau_A$ is the first hitting time of a set $A$, $\mathds{P}^{x,P}_{\pi}[.]$ denotes the state ($x$) dependent probability measure with a fixed policy $\pi^{x_0}$ for the MDP with transition probabilities $P(x,a,y)$. Similarly, we use $\mathds{E}_{\pi^{x_0}}^{x,\prob}[.]$ to denote the expectation with respect to transition measure $\mathds{P}^{x,P}_{\pi}[.]$. For a different MDP with transition probability denoted by $\hat{\prob}(x,a,y)$, expectation w.r.t. $\mathds{P}^{x,\hat{P}}_{\pi}[.]$ is denoted by $\mathds{E}_{\pi^{x_0}}^{x,\hat{\prob}}[.]$. If the expectation does not depend upon transition probability, we simply use $\mathds{E}_{\pi^{x_0}}^{x}[.]$.
 \par Following \cite{wisniewski2020p,wisniewski2023probabilistic}, we can express the safety function as below:
\begin{defn}[$p$-safety]\label{Def.psafe}
    For a given $p\in (0,1)$, we call a state $x$ to be $p$-safe (w.r.t. the policy $\pi^{x_0}$) if $S^P_{\pi^{x_0}}(x)\leq p$. 
\end{defn}
\par As given in \cite{wisniewski2023probabilistic}, one can relate the safety function with the transition probabilities to the forbidden states as
\begin{equation}
    S^P_{\pi^{x_0}}(x) = \mathds{E}_{\pi^{x_0}}^{x,\prob} \sum_{t = 0}^{\tau-1} \kappa(X_t, A_t),
    \label{safety_fn_exprsn}
\end{equation}
where $\tau := \tau_{E \cup U}$, and $\kappa(x,a) = \sum_{y \in U} \prob({x,a,y})$ for all $x \in \mathcal{X} \setminus V$. 
\begin{assum}
     For any policy $\pi$, the hitting time $\tau$ is finite almost surely, i.e., there exists a deterministic $ \tau_{max}<\infty$ such that $\mathds{P}^{x,P}_{\pi}[\tau \leq \tau_{max}]=1$.
 \end{assum}
   \begin{assum}
     An upper bound $\mathscr{T}$ on $\tau_{max}$ is known.
 \end{assum}
 \begin{assum}
     The process gets stopped whenever it reaches either the target set $E$ or the forbidden set $U$.
 \end{assum}
\par Suppose the taboo set $H$ of an MDP can be spitted into subsets $H\setminus U'$ and $U'$ such that only states in $U'$ can make a transition to the forbidden set $U$. Then, it is possible to be more exploratory on the set $H\setminus U'$ by applying any random policy while applying a safe baseline policy on the set $U'$. This motivates us to define a subset called \textit{proxy set}. 
\begin{defn}[Proxy Set]\label{Def.ProxySet}
Let $\pi$ be an arbitrary policy. Suppose that the set $U'\subset \mathcal{X}$. 
The subset $U'$ is a \textit{proxy set} if $\tau_{U'}<\tau_U$ almost surely. 
\end{defn}
\par The set $U'$ is a neighborhood of the forbidden set $U$ in the sense that the probability of hitting $U'$ before hitting $U$ is $1$.
\begin{rem}
     Consideration of the proxy set $U'$ is optional for the results presented in this paper. However, it can improve exploration. If no information is available on state-space to determine $U'$, the whole taboo $H$ is considered to be $U'$. Then, the setup becomes equivalent to the usual setup considered in the literature.
\end{rem}
\par To design a safe baseline policy for each state $x\in U'$, we need the notion of a safe action.
\begin{defn}[Safe Action]
    An action is called a safe action for a state $x\in H$, denoted by $a^{safe}(x)$, if $P(x,a^{safe}(x),y)=0$ for all $y\in U$.
\end{defn}
\par We now present the learning problem studied in the paper.\\
\\
\textbf{Problem $\mathcal{P}$:} 
\par We consider an unknown MDP with a given safety parameter $p$. For a given $x_0=\bar{x}\in H$, we find a policy $\pi^{*,\bar{x}}$, which is the solution of the following optimization problem:  
\begin{equation*}
\begin{split}
    &  \underset{\pi^{\bar{x}}\in \mathcal{D}^{\bar{x}}}{\text{argmax }} J_{\pi^{\bar{x}}}(\bar{x})\\
    & \text{such that } S^P_{\pi^{\bar{x}}}(\bar{x}) \leq p,
\end{split}
\end{equation*}
where
\begin{equation*}
   J_{\pi^{\bar{x}}}(\bar{x})=\mathds{E}^{\bar{x},P}_{\pi^{\bar{x}}} \sum_{t=0}^{\tau -1}\big[ r_t| X_0=\bar{x} \big]. 
\end{equation*}
\section{Safe policy Learning} \label{sec_Algo}
Before we present our algorithm, we provide some insights into probabilistic safety. 
\subsection{Probabilistic Safety:}
First, we demonstrate that the definition of probabilistic safety without stopping time is insufficient to ensure that the probability of visiting an unsafe state at any instant is less than a prescribed level. Without considering a stopping time or terminal states, probabilistic safety can be defined for a given instant \cite{el2018controlled}. 
To get the setup considered in \cite{el2018controlled}, we consider an MDP with a taboo set $H$ and an unsafe set $U$ that is terminal. There is no target set $E$. The notion of safety considered in \cite{el2018controlled} is the following. For a given policy $\pi$ and safety parameter vector $d=[d_1,...,d_{|\mathcal{X}|}]^T$, at instant $t$, the safety constraint is $\mathscr{P}_t\leq d$, where $\mathscr{P}_t=[\mathds{P}_{\pi}(X_t=1),...,\mathds{P}_{\pi}(X_t=|\mathcal{X}|)]^T$ and $\leq$ refers to element wise inequalities. 
\begin{lem} \label{safety_other}
   Suppose, for a given policy $\pi$ and safety parameter vector $d$, at instant $t$, the safety constraint $\mathscr{P}_t\leq d$ is satisfied. Then, with probability $1$, an unsafe state will be eventually visited.
\end{lem}
\begin{proof}
   Without loss of generality, suppose state $1$ is the unsafe state. We can always group all the unsafe states to form a new unsafe state. Since at instant $t$, $\mathscr{P}_t\leq d$, then, for a given $d_1\in (0,1)$,  $b:=\mathds{P}_{\pi}(X_t=1)\in (0,d_1)$. Now, if $X_0\notin U$ and the set $U$ is a terminal state, then at instant $t$,
   \small
   \begin{equation*}
       \begin{split}
           & \mathds{P}_{\pi}(X_t\in U) \\
            = & \mathds{P}_{\pi}(X_t\in U| X_{t-1}\notin U)\mathds{P}_{\pi}(X_t\notin U) \\
            = & (1-b)^{t-1}b
       \end{split}
   \end{equation*}
   \normalsize
   Therefore the probability that the unsafe state is visited at any instant $t$ is given by
\small
\begin{align*}
    \sum_{t=1}^{\infty} \mathds{P}_{\pi}(X_t\in U) = \sum_{t=1}^{\infty} (1-b)^{t-1}b = 1  
\end{align*}
\normalsize
\end{proof}
\par We now derive a recursive expression for the safety function. This expression is needed to prove that the proposed algorithm maintains safety in all episodes. Deriving a recursive expression for a value function is trivial for deterministic finite horizon and infinite horizon, as the expectation and the summation can commute in expression \eqref{safety_fn_exprsn}. However, with stochastic stopping time, it is not possible.  
\begin{lem} \label{safety_dp}
   For any $x\in H$, safety function can be written down in a recursive manner as follows:
    \begin{equation*}
        S^P_{\pi^{\bar{x}}}(x) = \mathds{E}^{x}_{\pi^{\bar{x}}} \Big[\kappa(x,a) + \sum_{y} \prob(x,a,y) S^P_{\pi^{\bar{x}}}(y) \big| a \sim \pi^{\bar{x}} \Big]
    \end{equation*}
\end{lem}
\begin{proof}
Note that $\mathds{E}_{\pi^{\bar{x}}}^{x,P}[\kappa(X_{t},A_{t})|\tau= l]= 0$ for all $t\geq l$. Now, we derive the following
\small
    \begin{equation*}
        \begin{split}
            &S^P_{\pi^{\bar{x}}}(x) \\
            &= \sum_{l=1}^{\infty}  \mathds{E}_{\pi^{\bar{x}}}^{x,P} \Big[ \sum_{t=0}^{l-1} \kappa(X_t,A_t)\Big| \tau = l \Big]\mathds{P}_{\pi^{\bar{x}}}^{x}[\tau = l]\\ 
           & = \mathds{E}_{\pi^{\bar{x}}}^{x,\prob} \Big[ \kappa(X_0,A_0)| \tau =1 \Big] \mathds{P}_{\pi^{\bar{x}}}^{x}[\tau = 1] \\
           & + \mathds{E}_{\pi^{\bar{x}}}^{x,\prob} \Big[ \kappa(X_1,A_1)| \tau =2 \Big] \mathds{P}_{\pi^{\bar{x}}}^{x}[\tau = 2]\\
           &+\cdots \\
            & = \mathds{E}_{\pi^{\bar{x}}}^{x,\prob} \Big[ \kappa(X_0,A_0)+\kappa(X_1,A_1)+\cdots| \tau =1 \Big] \mathds{P}_{\pi^{\bar{x}}}^{x}[\tau = 1] \\
           & + \mathds{E}_{\pi^{\bar{x}}}^{x,\prob} \Big[\kappa(X_0,A_0)+ \kappa(X_1,A_1)+\cdots | \tau =2 \Big] \mathds{P}_{\pi^{\bar{x}}}^{x}[\tau = 2]+\cdots \\
            & = \sum_{l=1}^{\infty} \mathds{E}_{\pi^{\bar{x}}}^{x,\prob} \Big[ \kappa(X_0,A_0)|\tau = l \Big] \mathds{P}_{\pi^{\bar{x}}}^{x}[\tau=l]\\ 
           & + \sum_{l=1}^{\infty} \mathds{E}_{\pi^{\bar{x}}}^{x,\prob} \Big[ \kappa(X_1,A_1)|\tau = l \Big] \mathds{P}_{\pi^{\bar{x}}}^{x}[\tau=l] +\cdots \\
           & = \mathds{E}_{\pi^{\bar{x}}}^{x,\prob}\Big[\kappa(X_0,A_0)\Big]+\mathds{E}_{\pi^{\bar{x}}}^{x,\prob}\Big[\kappa(X_1,A_1)\Big]+\cdots\\
            & =  \sum_{a\in A} \pi^{\bar{x}}(a|x) \kappa(x,a) \\
    & + \sum_{a\in A} \pi^{\bar{x}}(a|x) \sum_{y \in U'} \prob(x,a,y)  \sum_{a\in A} \pi^{\bar{x}}(a|y)  \kappa(y,a) +   \cdots \\
    &= \mathds{E}_{\pi^{\bar{x}}}^{x,\prob} \Big[\kappa(x,a) + \sum_{y} \prob(x,a,y) S^P_{\pi^{\bar{x}}}(y) \big| a \sim \pi^{\bar{x}} \Big]\\
     &= \mathds{E}_{\pi^{\bar{x}}}^{x} \Big[\kappa(x,a) + \sum_{y} \prob(x,a,y) S^P_{\pi^{\bar{x}}}(y) \big| a \sim \pi^{\bar{x}} \Big]
        \end{split}
    \end{equation*}
    \normalsize
     The infinite sum in the second equality is well defined. It is due to the fact that the taboo set consists of transient states. Then there exists a $\tau_{max}<\infty$ such that $\mathds{P}_{\pi^{\bar{x}}}^{x,P}[\tau>\tau_{max}]=0$. The last equality follows from the fact that the expectation is independent of transition probability $\prob$.
\end{proof}
\subsection{Safe Policy Design:}
\par If the transition probabilities are known, safe policies can be designed by solving the linear program given in Lemma \ref{PropositionAlgorithmSafeOpt}. The proof follows the same line of arguments as given in Proposition $2$ in \cite{wisniewski2023probabilistic}. To this end, for a fixed initial state $\bar{x}$, we define (state-action) occupation measure as follows:
\begin{equation*}
\begin{split}
 \gamma^{\bar{x}}(x,a) = \sum_{t=0}^{\infty} \mathds{P}_{\pi^{\bar{x}}}[X_t=x,A_t=a,t < \tau| x_0=\bar{x}]
\end{split}
\end{equation*} 
\begin{lem} \label{PropositionAlgorithmSafeOpt} 
For a fixed initial state $x_0=\bar{x}$, the optimal policy that solves Problem $\mathcal{P}$ is given by:
\begin{align}\label{PolicyGenerationConstraints}
  \pi^{*,{\bar{x}}}(a|x)  = \frac{{\gamma}^{*,{\bar{x}}}(x,a)}{\sum_{a \in \mathcal{A}} {\gamma}^{*,{\bar{x}}}(x,a)}.  
\end{align}
where, $\gamma^{*,{\bar{x}}}(x,a)$, $\forall(x,a)\in H \times \mathcal{U}$, are the solution of the following LP:
\small
\begin{equation}
 \max_{\gamma^{\bar{x}}} \sum_{x \in H} \sum_{a \in  \mathcal{A}} {\gamma^{\bar{x}}}(x,a) r(x,a) 
\label{LP1}
\end{equation}
\normalsize
subject to 
\small
\begin{equation}
\begin{split}
     & {\gamma^{\bar{x}}}(x, a) \geq 0, \hbox{ for } x \in V;\\
       & \delta_y(\bar{x}) + \sum_{a \in \mathcal{A}} \sum_{x \in H}{\gamma^{\bar{x}}}((x,a) (\prob({(x,a,y}) - \delta_y(x))= 0, \hbox{ for } y \in H;\\
      & \sum_{(x,a) \in H \times \mathcal{U}} {\gamma^{\bar{x}}}(x,a) \kappa(x,a)\leq p .
     \end{split}
\label{LP0}
\end{equation}
\normalsize
\end{lem}
\begin{rem}
   The setup considered in \cite{wisniewski2023probabilistic} is slightly different than the one considered here. We consider that the process is stopped whenever the forbidden set $U$ or the target set $E$ is reached for the first time. However, in \cite{wisniewski2023probabilistic}, it is possible to come out of the forbidden set $U$ and visit the target set $E$. Thus, two occupation measures are used therein as opposed to only one used in this paper. 
\end{rem}
\par As we deal with a model-free setting, we have to find the optimal and safe policy only using the estimation of $\prob(x,a,y)$. In the following subsection, we present an algorithm ( Algorithm $1$) in order to compute the safe optimal policy without any safety violation during the learning phase. 
\subsection{$p$-Safe Learning Algorithm (Algorithm $1$):} 
As the MDP has a finite stopping time, i.e., $\tau:=\tau_{U\cup E}<\infty$, our proposed algorithm (Algorithm $1$) works in episodes. 
Each episode starts with an initialization to a given initial state $\bar{x}$ and terminates at either a forbidden state or a goal state at a random time step. 
Since the transition probabilities are not known, at each episode $k$, we estimate the transition probabilities as follows: 
\begin{equation}
\begin{split}
& \forall x\in H, \textbf{ } \forall a\in A, \text{ }\forall y\in \mathcal{X}\\
    &\hat{\prob}_k({x,a,y}) = \frac{{N}_k(x,a,y)}{N_{k}(x,a) \vee 1}
\end{split}
\label{sample_avg}
\end{equation}
 where, $N_k(x,a)$ and $N_k(x,a,y)$ the number of visits to $(x,a)$ and $(x,a,y)$, respectively, before iteration $k$. 
\par One can use the estimated transition probabilities $\hat{\prob}_k(x,a,y)$ and estimated $\hat{\kappa}_k(x,a):= \sum_{y\in U} \hat{\prob}_k(x,a,y)$ in LP \eqref{LP1}-\eqref{LP0} to get an estimated optimal policy. However, the feasibility of the LP in all episodes is not guaranteed. Further, even if LP is feasible, safety is not guaranteed. To circumvent the above challenges, our proposed algorithm will be based on the OFU idea. As OFU alone does not guarantee satisfaction with the safety constraint, we modify the safety constraints by considering more conservative constraints. To find the p-safe policy, our algorithm solves the following optimization problem:
\small
\begin{equation}
 \begin{split}
     &\underset{\tilde{P}\in G_k,\hat{\gamma}^{\bar{x}}}{\text{max}} \sum_{(x,a)\in H\times \mathcal{A}} \hat{\gamma}^{\bar{x}}(x,a) (r(x,a) + \hat{\epsilon}_k(x,a) ) \\
     & \text{s.t. } \hat{\gamma}^{\bar{x}}(x, a) \geq 0, \hbox{ for } (x,a) \in H\times \mathcal{A};\\
       & \delta_y(x_0) + \sum_{a \in \mathcal{A}} \sum_{x \in H}\hat{\gamma}^{\bar{x}}((x,a) (\tilde{\prob}({(x,a,y}) - \delta_y(x))= 0, \ y \in H; \\
     & \sum_{(x,a)\in H\times \mathcal{A}} \hat{\gamma}^{\bar{x}}(x,a) (\hat{\kappa}_k(x,a)+ 3\hat{\epsilon}_k(x,a))\leq p,
 \end{split}   
 \label{ofu_pol}
\end{equation}
\normalsize
where, $G_k=\{g_k(x,a) \ | \sum_{y\in \mathcal{X}} \tilde{\prob}(x,a,y)=1 ; \forall (x,a)\in \mathcal{X}\times \mathcal{A} \}$, $g_k(x,a)=\{\tilde{\prob}(x,a,y): |\tilde{\prob}(x,a,y)-\hat{\prob}(x,a,y)|\leq \epsilon_k(x,a,y); \forall y\in \mathcal{X} \}$, $\hat{\gamma}^{\bar{x}}(x,a)$ is the state-action occupation measure with transition probability $\tilde{P}$, $\hat{\epsilon}_k(x,a)=\sum_{y\in \mathcal{X}} \epsilon_k(x,a,y)$, and $\epsilon_k(x,a,y)$ is as given below, 
\small
\begin{equation*}
    \begin{split}
       & \epsilon_k(x,a,y)\\
       &= \sqrt{\frac{4\hat{\prob}_k(x,a,y)(1-\hat{\prob}_k(x,a,y))\text{log}(\frac{2|\mathcal{X}||\mathcal{A}|K}{w})}{N_k(x,a)\vee 1}} \\
       & + \frac{14\text{ log}\big(\frac{2|\mathcal{X}||\mathcal{A}|K}{w}\big)}{3((N_k(x,a)-1)\vee 1)}  
    \end{split}
\end{equation*}
 \normalsize
The optimal policy of the above optimization problem is given by $\tilde{\pi}^{\bar{x}}_k(a|x)=\frac{\hat{\gamma}^{\bar{x}}(x,a)}{{\sum_{y\in H}\hat{\gamma}^{\bar{x}}}(y,a)}$. During the episode $k$, policy obtained $\tilde{\pi}^{\bar{x}}_k$ is applied until the terminal states are reached, and the counters $N_k(x,a)$ and $N_k(x,a,y)$ are updated in each transition. Whenever, the episode is ended, a new episode starts from the given initial state $\bar{x}$. Note that, to incentivize exploration of less visited state-action pairs, we add the term $\hat{\epsilon}_k(x,a)$ to the corresponding reward $r(x,a)$. Since $\tilde{P}_k(x,a,y) = P(x,a,y)$ and $\epsilon_k(x,a,y)=0$ in the limit as $k$ increases, optimization \eqref{ofu_pol} becomes equivalent to the LP given in Lemma \ref{PropositionAlgorithmSafeOpt}, hence the policy $\tilde{\pi}^{\bar{x}}_k$ converges to the optimal policy $\pi^{*,\bar{x}}$. We call $\tilde{P}$ as the optimistic transition probability as it maximizes optimization \eqref{ofu_pol}.   
\par We now present a  result that relates the true transition probability $P$ with the estimated probability $\hat{P}$ and the optimistic probability $\tilde{P}$.
 \begin{lem}\label{concen_lem}
    Consider $w\in (0,1)$. Now, with a probability at least $(1-2w)$, for all $(x,a,y)\in {H}\times \mathcal{A}\times \mathcal{\mathcal{X}}$ and $k\in \{1,2,...,K\}$, the following is true:
    \begin{enumerate}[i)]
    \item \small
 \begin{equation*}
 \begin{split}
   \text{  }  |\prob(x,a,y) - \hat{\prob}_k(x,a,y)|\leq \epsilon_k(x,a,y), 
 \end{split}
 \end{equation*}
 \normalsize 
    \item \small
    \begin{equation*}
        \begin{split}
            |\prob(x,a,y)-\tilde{\prob}(x,a,y)|\leq 2\epsilon_k(x,a,y).
        \end{split}
    \end{equation*}
    \normalsize
    \item \small
    \begin{equation*}
        \kappa(x,a) \leq \hat{\kappa}_k(x,a) + \hat{\epsilon}_k(x,a)
    \end{equation*}
    \normalsize
    \end{enumerate}
 \end{lem}
\begin{proof} Proof of claim i):
    Using the empirical Bernstein inequality  (Theorem $4$ in \cite{maurer2009empirical}), for fixed $(x,a,y)$, episode $k$ and $\eta$ such that $2\eta\in (0,1)$, we get
    \small
    \begin{equation*}
     \begin{split}
         \mathds{P}\Big[|\prob(x,a,y)-\hat{\prob}_k(x,a,y)|\geq \alpha_k(x,a,y)\Big] \leq 2\eta,
     \end{split}   
    \end{equation*}
    \normalsize
    where 
    \small
    \begin{equation*}
      \begin{split}
       & \alpha_k(x,a,y)\\
       &= \sqrt{\frac{2\hat{\prob}_k(x,a,y)(1-\hat{\prob}_k(x,a,y))\text{log}(\frac{2}{\eta}))}{N_k(x,a)\vee 1}} \\
       & + \frac{7\text{ log}\big(\frac{2}{\eta}\big)}{3((N_k(x,a)-1)\vee 1)}  
    \end{split}  
    \end{equation*}
    \normalsize
    Now, set $\eta=\frac{w}{|\mathcal{X}|^2|\mathcal{A}|K}$. Then, considering all the combinations of $(x,a,y)$ with all episodes up to $K$ and using union bound, we get the following for any $(x,a,y)\in H\times \mathcal{A}\times \mathcal{X}$ and any episode $k$:
    \small
 \begin{equation*}
 \begin{split}
  \mathds{P}\Big[ | \prob(x,a,y) - \hat{\prob}_k(x,a,y) | \geq \xi_k(x,a,y)\Big] \leq 2w, 
 \end{split}
 \end{equation*}
 \normalsize 
 where
 \small
\begin{equation*}
    \begin{split}
       & \xi_k(x,a,y)\\
       &= \sqrt{\frac{2\hat{\prob}_k(x,a,y)(1-\hat{\prob}_k(x,a,y))\text{log}(\frac{2|\mathcal{X}|^2|\mathcal{A}|K}{w}))}{N_k(x,a)\vee 1}} \\
       & + \frac{7\text{ log}\big(\frac{2|\mathcal{X}|^2|\mathcal{A}|K}{w}\big)}{3((N_k(x,a)-1)\vee 1)}  \\
       & \leq \sqrt{\frac{2\hat{\prob}_k(x,a,y)(1-\hat{\prob}_k(x,a,y))\text{log}\big((\frac{2|\mathcal{X}||\mathcal{A}|K}{w})^2\big)}{N_k(x,a)\vee 1}} \\
       & + \frac{7\text{ log}\big((\frac{2|\mathcal{X}||\mathcal{A}|K}{w})^2\big)}{3((N_k(x,a)-1)\vee 1)}  \\
       &= \sqrt{\frac{4\hat{\prob}_k(x,a,y)(1-\hat{\prob}_k(x,a,y))\text{log}(\frac{2|\mathcal{X}||\mathcal{A}|K}{w}))}{N_k(x,a)\vee 1}} \\
       & + \frac{14\text{ log}\big(\frac{2|\mathcal{X}||\mathcal{A}|K}{w}\big)}{3((N_k(x,a)-1)\vee 1)}.  
    \end{split}
\end{equation*}
 \normalsize
 Hence we get the desired result by taking  compliment. 
 \par Proof of claim ii): From claim i), we get that with probability at least $(1-2w)$, $|\tilde{\prob}(x,a,y)-\hat{\prob}(x,a,y)|\leq \epsilon_k(x,a,y)$ and $|{\prob}(x,a,y)-\hat{\prob}(x,a,y)|\leq \epsilon_k(x,a,y)$. Now using triangular inequality, we have the desired result.
 \par Proof of claim iii): From claim i), with probability at least $(1-2w)$, for all $(x,a,y)\in H\times \mathcal{A} \times \mathcal{X}$ and episode $k\in \{1,2,...,K\}$, we have the following:
    \small
    \begin{equation*}
    \begin{split}
       & \prob(x,a,y) \leq \hat{\prob}_k(x,a,y) + \epsilon_k(x,a,y) \\
        \implies & \sum_{y\in U} \prob(x,a,y) \leq \sum_{y\in U} [ \hat{\prob}_k(x,a,y) + \epsilon_k(x,a,y)]\\
        \implies & \kappa(x,a) \leq \hat{\kappa}_k(x,a) + \hat{\epsilon}_k(x,a)
    \end{split}
    \end{equation*}
    \normalsize
\end{proof}
 \par In the sequel, we show that if optimization \eqref{ofu_pol} is feasible, then the estimated policy is a safe policy with probability at least $(1-2w)$. The optimization problem \eqref{ofu_pol} can be solved efficiently by casting it as an \textit{extended linear program} \cite{efroni2020exploration}. To this end, we consider state-action-state occupation measure $\hat{\beta}^{\bar{x}}(x,a,y):={\prob}(x,a,y)\hat{\gamma}^{\bar{x}}(x,a)$. In each episode $k\in \{1,2,...,K\}$, with the estimated transition probabilities $\hat{\prob}_k$, we solve the following extended LP and obtain a policy.
\small
\begin{equation}
 \max_{\hat{\beta}^{\bar{x}}} \sum_{(x,a,y) \in H\times \mathcal{A}\times \mathcal{X}} \hat{\beta}^{\bar{x}}(x,a,y) (r(x,a) + \hat{\epsilon}_k(x,a) ) 
\label{LP1_est}
\end{equation}
\normalsize
s.t. 
\small
\begin{equation}
\begin{split}
    &  \hat{\beta}^{\bar{x}}(x, a,y)\geq 0, \hbox{ for } (x,a,y) \in H\times \mathcal{A}\times \mathcal{X}\\
     & \delta_y(\bar{x}) + \sum_{(x,a) \in H\times \mathcal{A}} \hat{\beta}^{\bar{x}}(x,a,y)  - \sum_{(a,z) \in  \mathcal{A}\times \mathcal{X}} \hat{\beta}^{\bar{x}}(y,a,z)=0, \\
     & \hspace{7cm} \forall y \in H;\\
     & \hat{\beta}^{\bar{x}}(x,a,y) - (\hat{\prob}_k(x,a,y)+\epsilon_k(x,a,y))\sum_{z\in \mathcal{X}} \hat{\beta}^{\bar{x}}(x,a,z) \leq 0, \\
     & \hspace{5cm} \forall (x,a,y)\in H\times \mathcal{A}\times \mathcal{X}; \\   
     & -\hat{\beta}^{\bar{x}}(x,a,y) + (\hat{\prob}_k(x,a,y)-\epsilon_k(x,a,y))\sum_{z\in \mathcal{X}} \hat{\beta}^{\bar{x}}(x,a,z) \leq 0, \\
     & \hspace{5cm} \forall (x,a,y)\in H\times \mathcal{A}\times \mathcal{X}\\
      & \sum_{(x,a,z) \in H \times \mathcal{U}\times \mathcal{X}}  \hat{\beta}^{\bar{x}}(x,a,z) (\hat{\kappa}_k(x,a)+ 3 \hat{\epsilon}_k(x,a))\leq p,
     \end{split}
\label{LP0_est}
\end{equation}
\normalsize
\par The policy at episode $k$ is given by 
\begin{equation}
 \tilde{\pi}^{\bar{x}}_k(a|x)=\frac{\sum_{z\in \mathcal{X}}\hat{\beta}^{\bar{x}}(x,a,z)}{\sum_{(a',z)\in \mathcal{A}\times \mathcal{X}}\hat{\beta}^{\bar{x}}(x,a',z)}
 \label{estimate_pol}
\end{equation} 
\par Until the extended LP \eqref{LP1_est}-\eqref{LP0_est} is feasible, we apply a safe baseline policy. If at any episode $k$, LP \eqref{LP1_est}-\eqref{LP0_est} is feasible, we prove that the policy $\tilde{\pi}^{\bar{x}}_k$ obtained using \eqref{estimate_pol} is a safe policy with a prescribed high probability. Thus, the safety constraint is not violated with high probability.  
 \\
 \\
 \textit{{Safe baseline policy design:}}
 \par Now, we develop baseline policy that is safe. While this baseline policy is not necessarily optimal, it is crucial for ensuring that the constraints are not violated during the policy learning. Further, for the safety of any state $x\in H$, it is sufficient to apply safe policy only at the proxy states and any arbitrary policy can be applied to other states. The following result states this. However, if proxy states are not given, we apply the safe baseline policy for all states in the taboo set $H$. 
 \begin{lem} \label{safety_proxy} \cite{mazumdar2023online}
     The safety function satisfies the following property:
     \begin{equation*}
         S^P_{\pi^{\bar{x}}}(x) \leq S^P_{\pi^{\bar{x}}}(x'); \ \forall x\in H\setminus U' \ \text{and} \ x'\in U'.
     \end{equation*}
 \end{lem}
\par In the following result, we design a safe baseline policy.
\begin{thm}\label{safe_pol}
    Suppose a safe action ${a}^{safe}(x')$ is known for each proxy state $x'\in U'$. Suppose $1\geq q\geq 1-\frac{p}{\mathscr{T}}$, where $\mathscr{T}\geq \tau_{max}$. Then the following policy $\pi^S$, that does not depend on the initial state $\bar{x}$, is a safe baseline policy for each state $x\in H$: 
     \begin{equation}
    \begin{split}
        \pi^S(a|x) = \begin{cases}
               \check{\pi}^S(a|x'), & \text{if } x'\in U' \\
               \text{random policy}, & \text{if } x'\in H \setminus U',
                \end{cases}
    \end{split}
\end{equation}
where, for all $x'\in U'$,
    \begin{equation}
    \begin{split}
        \check{\pi}^S(a|x') = \begin{cases}
               {q}, & \text{if } a= {a}^{safe}(x') \\
               \frac{(1-q)}{|\mathcal{A}|-1}, & \text{otherwise},
                \end{cases}
    \end{split}
\end{equation}
\end{thm}
\begin{proof}
The safety function for each state $x'\in U'$ with the policy ${\pi}^S$ can be expressed as follows. 
    \small
    \begin{equation}
        \begin{split}
           & S^P_{\pi^S}(x') \\
           & = \mathds{E}_{\pi^S}^{x',P} [\sum_{t=0}^{\tau-1} \kappa(X_t,A_t)]\\
           & \overset{(a)}{=} \mathds{P}_{\pi^S}^{x'} [\tau\leq \tau_{max}] \mathds{E}_{\pi^S}^{x',P} [\sum_{t=0}^{\tau_{max}-1} \kappa(X_t,A_t)]\\
           & + \mathds{P}_{\pi^S}^{x'} [\tau> \tau_{max}] \mathds{E}_{\pi^S}^{x',P} [\sum_{t=\tau_{max}}^{\infty} \kappa(X_t,A_t)] \\
           & \overset{(b)}{=} \mathds{E}_{\pi^S}^{x',P} [\sum_{t=0}^{\tau_{max}-1} \kappa(X_t,A_t)] \\
            & \overset{(c)}{=}  \sum_{a\in A} \pi^S(a|x') \kappa(x',a) \\
    & + \sum_{\tilde{x}' \in U'} \mathds{P}^{x'}_{\pi^S} [X_1=\tilde{x}']  \sum_{a'\in A} \pi^S(a|\tilde{x}')  \kappa(\tilde{x}',a') +   \cdots \\
    & +  \sum_{\tilde{x}'\in U'} \mathds{P}^{x'}_{\pi^S} [X_{\tau_{max}-1}=\tilde{x}'] \\
    & \cdot \sum_{a'\in A} \pi^S(a|\tilde{x}')  \kappa(\tilde{x}',a') \\    
   & \overset{(d)}{\leq} \tau_{max}  \underset{\tilde{x}'\in U'}{\text{max}} \sum_{a\in A} \pi^S(a|\tilde{x}') \kappa(\tilde{x}',a) \\
    & {\leq}  \mathcal{T} \underset{\tilde{x}'\in U'}{\text{max}} \sum_{a\in A} \pi^S(a|\tilde{x}') \kappa(\tilde{x}',a) \\
           & = \mathcal{T} \underset{\tilde{x}'\in U'}{\text{max}} \sum_{y\in U} \Big[ q \cdot \prob(\tilde{x}',a^{safe}(\tilde{x}'),y) \\ 
            & + \sum_{a\in A\setminus \{a^{safe}(x')\}}  \frac{(1-q)}{|\mathcal{A}|-1} \cdot \prob(\tilde{x}',a,y) \Big] \\
           & \overset{(e)}{=} \mathcal{T} \underset{\tilde{x}'\in U'}{\text{max}}  \sum_{a\in A\setminus \{a^{safe}(x')\}} \frac{(1-q)}{|\mathcal{A}|-1} \cdot \sum_{y\in U} \prob(x',a,y) \\
           & \overset{(f)}{\leq} \mathcal{T} \frac{(|\mathcal{A}|-1)}{(|\mathcal{A}|-1)} (1-q) \\
           & = \mathcal{T} (1-q)\\
           & \overset{(g)}{\leq} p
        \end{split}
    \end{equation}
    \normalsize
    Equality $(a)$ takes care of all the possible realizations. Equality $(b)$ follows as $\mathds{P}_{\pi^S}^{x'} [\tau\leq \tau_{max}]=1$. We get equality $(c)$ using Lemma \ref{safety_dp}. Inequalities $(d)$ and $(f)$ follow as the probability is upper bounded by $1$. Equality $(e)$ follows from the fact that if a safe action is selected, i.e., $a=a^{safe}(x')$, then the process does not visit any forbidden state, almost surely, i.e., $\prob(x',a^{safe}(x'),y)=0$ for all $y\in U$. Finally, we get inequality $(g)$ from the definition of $q$. Thus, each proxy state $x'\in U'$ is safe. Hence, in view of Lemma 1, any initial state $x_0\in H$ is safe.
\end{proof}
\par In the sequel, we use the following notations:
\small
\begin{equation*}
\begin{split}
& \hat{S}^{\tilde{P}}_{\tilde{\pi}^{\bar{x}}_k}(\bar{x}) = \mathds{E}_{\hat{\pi}^{\bar{x}}_k}^{\tilde{P}}\Big[\sum_{t=0}^{\tau-1}   \hat{\kappa}_k(X_t,A_t)  |X_0=\bar{x} \Big] \\
&    \tilde{S}^{\tilde{P}}_{\tilde{\pi}^{\bar{x}}_k}(\bar{x}) = \mathds{E}_{\hat{\pi}^{\bar{x}}_k}^{\tilde{P}}\Big[\sum_{t=0}^{\tau-1}   \hat{\kappa}_k(X_t,A_t)  + 3  \hat{\epsilon}_k(X_t,A_t) |X_0=\bar{x} \Big]
    \end{split}
\end{equation*}
\normalsize
\par We now state one of the main results of the paper. In the following result, we show that the policy $\tilde{\pi}_k^{\bar{x}}$ is a safe policy with an arbitrary high probability.
\begin{thm} \label{Safety_proof}
With probability at least $(1-2w)$, the policy $\tilde{\pi}_k$ designed using Algorithm $1$ is safe, i.e.,
\begin{align*}
    S^P_{\tilde{\pi}^{\bar{x}}_k}(\bar{x}) \leq p.
\end{align*} 
\end{thm}
\begin{proof}
If the LP \eqref{LP1_est}-\eqref{LP0_est} is feasible then, with the estimated optimistic transition probability $\prob(x,a,y)$, $\tilde{S}^{\tilde{P}}_{\tilde{\pi}^{\bar{x}}_k}(\bar{x})\leq p$.
Now, we show that $S^P_{\tilde{\pi}^{\bar{x}}_k}(\bar{x}) \leq p$. With the policy $\tilde{\pi}^{\bar{x}}_k$, consider the following: 
\small
\begin{equation*}
    \begin{split}
         &S^P_{\tilde{\pi}^{\bar{x}}_k}(\bar{x}) - \hat{S}^{\tilde{P}}_{\tilde{\pi}^{\bar{x}}_k}(\bar{x}) \\
         & = S^P_{\tilde{\pi}^{\bar{x}}_k}(\bar{x}) \\
         &-\mathds{E}_{\tilde{\pi}^{\bar{x}}_k} [\hat{\kappa}(X_0,A_0)+ \sum_{X_{1}\in H} \tilde{\prob}_k(X_0,A_0,X_1) {S}^P_{\tilde{\pi}^{\bar{x}}_k}(X_{1})|X_0=\bar{x}]\\
         &+ {E}_{\tilde{\pi}^{\bar{x}}_k} [\hat{\kappa}(X_0,A_0)+ \sum_{X_{1}\in H} \tilde{\prob}_k(X_0,A_0,X_{1}) {S}^P_{\tilde{\pi}^{\bar{x}}_k}(X_{1})|X_0=\bar{x}] \\
         &- \hat{S}^{\tilde{P}}_{\tilde{\pi}^{\bar{x}}_k}(X_0)  \\
        & =  \mathds{E}_{\tilde{\pi}^{\bar{x}}_k} [(\kappa(X_0,A_0)-\hat{\kappa}(X_0,A_0))\\
        &+ \sum_{X_{1}\in H} (\prob(X_0,A_0,X_{1})-\tilde{\prob}^{\bar{x}}_k(X_0,A_0,X_{1})) {S}^P_{\tilde{\pi}^{\bar{x}}_k}(X_{1})|X_0=\bar{x}]\\
        & + \mathds{E}_{\tilde{\pi}^{\bar{x}}_k} [ \sum_{X_{1}\in H} \tilde{\prob}_k(X_0,A_0,X_{1}) ({S}^P_{\tilde{\pi}^{\bar{x}}_k}(X_{1})-\hat{S}^{\tilde{P}}_{\tilde{\pi}^{\bar{x}}_k}(X_{1}))|X_0=\bar{x}]
    \end{split}
\end{equation*}
\normalsize
Similarly,
\small
\begin{equation*}
    \begin{split}
       &S_{\tilde{\pi}^{\bar{x}}_k}(X_{1}) - \hat{S}^{\tilde{P}}_{\tilde{\pi}^{\bar{x}}_k}(X_{1}) \\
         & = \mathds{E}_{\tilde{\pi}^{\bar{x}}_k} [(\kappa(X_{1},A_{1})-\hat{\kappa}(X_{1},A_{1}))\\
        &+ \sum_{X_{2}\in H} (p(X_{1},A_{1},X_{2})-\tilde{\prob}_k(X_{1},A_{1},X_{2})) {S}^P_{\tilde{\pi}^{\bar{x}}_k}(X_{2})|X_{1}]\\
        & + \mathds{E}_{\tilde{\pi}^{\bar{x}}_k} [ \sum_{X_{2}\in H} \tilde{\prob}_k(X_{1},A_{1},X_{2}) ({S}_{\tilde{\pi}^{\bar{x}}_k}(X_{2})-\hat{S}^{\tilde{P}}_{\tilde{\pi}^{\bar{x}}_k}(X_{2}))|X_{1}]
    \end{split}
\end{equation*}
\normalsize
Using recursive expression, we get the following:
\small
\begin{equation*}
    \begin{split}
       &S_{\tilde{\pi}^{\bar{x}}_k}(\bar{x}) - \hat{S}^{\tilde{P}}_{\tilde{\pi}^{\bar{x}}_k}(\bar{x}) \\
         & = \mathds{E}_{\tilde{\pi}^{\bar{x}}_k}^{\hat{P}}\Big[\sum_{t=0}^{\tau-1} (\kappa(X_t,A_t)-\hat{\kappa}(X_t,A_t))\\
        &+ \sum_{y\in \mathcal{X}} (\prob(X_t,A_t,y)-\tilde{\prob}_k(X_t,A_t,y)) \hat{S}^{\tilde{P}}_{\tilde{\pi}^{\bar{x}}_k}(y)|X_0=\bar{x} \Big]\\
        & \overset{(a)}{\leq} \mathds{E}_{\tilde{\pi}^{\bar{x}}_k}^{\hat{P}}\Big[\sum_{t=0}^{\tau-1} (\kappa(X_t,A_t)-\hat{\kappa}(X_t,A_t))\\
        &+ \sum_{y\in \mathcal{X}} (\prob(X_t,A_t,y)-\tilde{\prob}_k(X_t,A_t,y)) |X_0=\bar{x} \Big]\\
        & \overset{(b)}{\leq} \mathds{E}_{\tilde{\pi}^{\bar{x}}_k}^{\hat{\prob}}\Big[\sum_{t=0}^{\tau-1} 3  \hat{\epsilon}_k(X_t,A_t) |X_0=\bar{x} \Big]
\end{split}
\end{equation*}
\normalsize
Inequality $(a)$ follows as the safety function is upper bounded by $1$ and $(b)$ is follows from Lemma \ref{concen_lem}. Hence, 
\small
\begin{equation}
    \begin{split}
     & S^P_{\tilde{\pi}^{\bar{x}}_k}(\bar{x}) \\
     & \leq \hat{S}^{\tilde{P}}_{\tilde{\pi}^{\bar{x}}_k}(\bar{x}) + \mathds{E}_{\tilde{\pi}^{\bar{x}}_k}^{\hat{p}}\Big[\sum_{t=0}^{\tau-1} 3  \hat{\epsilon}_k(X_t,A_t) |X_0=\bar{x} \Big] \\
     & = \mathds{E}_{\tilde{\pi}^{\bar{x}}_k}^{\hat{P}}\Big[\sum_{t=0}^{\tau-1}   \hat{\kappa}_k(X_t,A_t)  + 3  \hat{\epsilon}_k(X_t,A_t) |X_0=\bar{x} \Big]\\
    & = \tilde{S}^{\tilde{P}}_{\tilde{\pi}^{\bar{x}}_k}(\bar{x})\\
    & \leq p
    \end{split}
    \label{safe_ineq}
\end{equation}
\normalsize
Therefore, with probability at least $(1-2w)$, policy $\tilde{\pi}_k^{\bar{x}}$ is safe policy. 
\end{proof}
\small
\begin{algorithm} \label{safe_algo}
  \caption{: $p$-safe RL algorithm:}
  \begin{algorithmic}[1]
    \State \textbf{Input:} Safety parameter $p\in (0,1)$, initial state $\bar{x}$, confidence parameter $w\in (0,1)$, at least one safe action $a^{safe}$ for each state, reward $r(x,a)$ for all $(x,a)\in H\times \mathcal{A}$
    \State \textbf{Initialize:} Select $N_{1}(x,a)=0$, $N_{1}(x,a,y)=0$ for all $(x,a,y)\in H\times \mathcal{A}\times \mathcal{X}$
    \State Set $l=1$
    \For  {Episodes ($k=1,2,...,K$)}
    \State Set $X_0 \leftarrow \bar{x}$
    \State Set $N_k(x,a)\leftarrow N_l(x,a)$ and $N_k(x,a,y)\leftarrow N_l(x,a,y)$ for all $x,a,y$
     \State Compute $\hat{\prob}_k(x,a,y)$ using Equation \eqref{sample_avg}
    \State Solve LP \eqref{LP1_est}-\eqref{LP0_est} using the estimate $\hat{\prob}_k(x,a,y)$
    \If{the LP is feasible}
    \State Obtain the policy $\tilde{\pi}^{\bar{x}}_{k}$ using \eqref{estimate_pol}
    \Else
    \State Set the policy $\tilde{\pi}^{\bar{x}}_k$ as the safe baseline policy $\pi^S$
    \EndIf
  \For  { ($t=0,2,...,T$)}
    \If  { $X_t$ is not a terminal state}
    \State  Apply action $A_0$ according to the policy $\tilde{\pi}^{\bar{x}}_{k}(X_t)$
    \State Observe the next state $X_{t+1}$ and the reward $r_t$ 
    \State Update counters $N_{l+1}(X_t,A_t) \leftarrow N_{l}(X_t,A_t)+1$  and $N_{l+1}(X_t,A_t,X_{t+1}) \leftarrow N_{l}(X_t,A_t,X_{t+1})+ 1$
      \Else 
     \State Terminate the loop
    \EndIf
    \State Set $l \leftarrow l+1$
     \EndFor
     \EndFor
  \end{algorithmic}
\end{algorithm}
\normalsize
 \section{Illustrating Example} \label{example}
\begin{figure}[!htb] 
  \centering
  \includegraphics[scale=0.6]{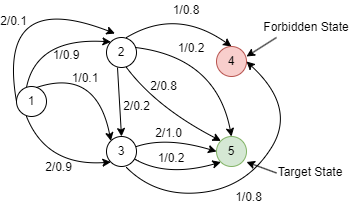}
\caption{Example MDP}
\label{fig:CDC_MDP}
\end{figure}

\begin{figure}[!htb] 
  \includegraphics[scale=0.35]{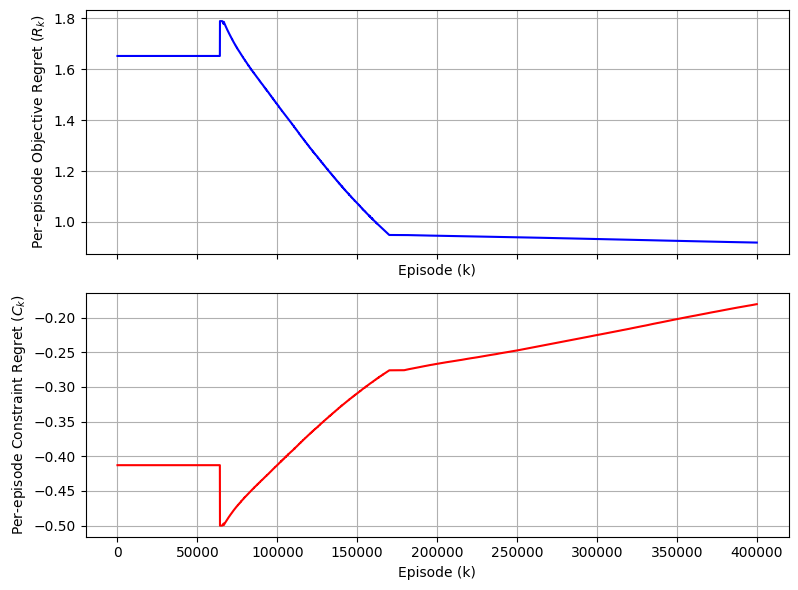}
\caption{ Evolution of Per-episode Objective and Constraint Regret}
\label{fig:rr_avg}
\end{figure}

\begin{figure}[!htb] 
  \includegraphics[scale=0.47]{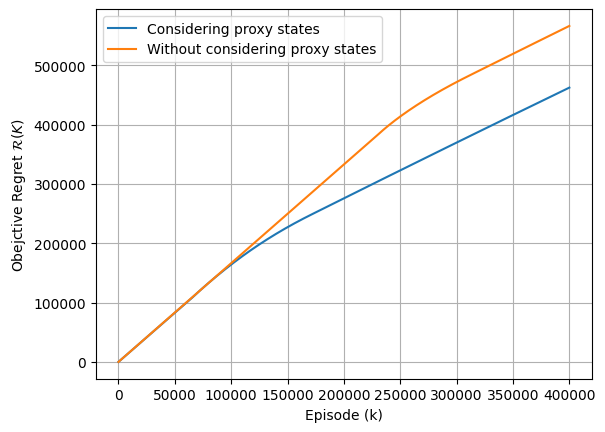}
\caption{ Objective Regret with and without knowledge of the proxy states}
\label{fig:rr_prox_vs_no_prox_avg}
\end{figure}

  Consider the MDP as shown in Figure \ref{fig:CDC_MDP}. There are $5$ states given by $\mathcal{X}=\{1,2,3,4,5\}$ and two actions given by $\mathcal{A}=\{1,2\}$. The taboo set is $H=\{1,2,3\}$, Forbidden state is $U=4$ and the target state is $E=5$. Clearly, the proxy set is $\{2,3\}$. State transition probabilities are as follows: $P(1,1,2)= 0.9$, $P(1,1,3)= 0.1$, $P(1,2,2)= 0.1$,  $P(1,2,3)= 0.9$, $P(2,1,4)=0.8$, $P(2,1,5)=0.2$, $P(2,2,3)=0.2$, $P(2,2,5)=0.8$, $P(3,1,4)= 0.8$, $P(3,1,5)= 0.2$, $P(3,2,5)=1$. The remaining transition probabilities are $0$. The confidence parameter is considered to be $w= 0.01$. Reward associated with each state-action pair is as follows: $r(1,1)=1$, $r(1,2)=1$, $r(2,1)=2$, $r(2,2)=1$, $r(3,1)=4$ and $r(3,2)=1$.
To assess the effectiveness of the algorithm, we calculate the per-episode objective regret $R_k$ and per-episode constraint regret $C_k$ defined below:
\small
\begin{equation*}
 R_{k} = J_{\pi^{*,{\bar{x}}}}(\bar{x}) - J_{\tilde{\pi}^{\bar{x}}_k}(\bar{x}); \     C_{k} = S^P_{\tilde{\pi}^{\bar{x}}_k}(\bar{x}) - p. 
\end{equation*}
\normalsize
We also evaluate the objective regret defined as 
\small
\begin{equation*}
  \mathcal{R}(K) = \sum_{k=1}^{K} R_k
\end{equation*}
\normalsize
With the given transition policy, the optimal policy computed using Lemma \ref{PropositionAlgorithmSafeOpt} is $\pi^{*,{\bar{x}}}(1|1)=0.4609375$, $\pi^{*,{\bar{x}}}(2|1)=0.5390625$, $\pi^{*,{\bar{x}}}(1|2)=0$, $\pi^{*,{\bar{x}}}(2|2)=1.0$, $\pi^{*,{\bar{x}}}(1|3)=1.0$ and $\pi^{*,{\bar{x}}}(2|3)=0$. We consider state $1$ to be the initial state, i.e., $\bar{x}=1$ and the safety parameter $p=0.5$. The optimal cost is $J_{\pi^{*,{\bar{x}}}}(\bar{x})=3.96875$. Considering action $2$ to be the safe action for each proxy state $x'\in \{2,3\}$, we compute a safe baseline policy for the proxy states using Theorem \ref{safe_pol}. To this end we consider $\mathcal{T}=5\geq \tau_{max}$. A safe baseline policy is $\check{\pi}^S(1|x')=0.9\geq 1- \frac{0.5}{\mathcal{T}}$ and $\check{\pi}^S(1|x')=0.1$. In Fig. \ref{fig:rr_avg}, we show the per-episode objective regret $R_k$ and constraint regret $C_k$ achieved with a policy learned using Algorithm $1$. As desired, we see a reducing per-episode objective regret as the number of episode increases. Hence, a sublinear objective regret is achieved. Further, the per-episode constraint regret is always less than $0$, implying satisfaction of the safety constraint in all episodes.   
In Fig. \ref{fig:rr_prox_vs_no_prox_avg}, we illustrate the objective regret considering both proxy states and without considering any proxy states. For the case with proxy states, we apply the safe baseline policy only at the proxy states, i.e. $\{2,3\}$ and apply an uniform policy at state $1$. However, when we do not assume any knowledge about the proxy states, we apply the safe baseline policy at all the states in the taboo set $H=\{1,2,3\}$. From Fig. \ref{fig:rr_prox_vs_no_prox_avg}, we see that knowledge about the proxy states results in achieving better  objective regret than not having knowledge. 
\section{Conclusion and Future Work} 
\label{sec_conclusion}
In this paper, we have presented an online algorithm for learning the optimal policy for an MDP without any safety violation. Our algorithm is based upon \textit{optimism in the face of uncertainty} principle. Since, during initial episodes, the estimated policy could be unsafe, the developed algorithm applies a safe baseline policy. To this end, we introduced a method for designing a safe baseline policy. We have demonstrated that a subset of the state-space referred to as \textit{proxy set} being known leads to better regret than not having any information. Further, we demonstrate that if stopping time is not incorporated into the definition of probabilistic safety, the notion of safety becomes inconsistent.
\par As future work, we will derive regret bounds for the proposed algorithm. Further, another extension will be to consider a large state space with function approximation. 

\addtolength{\textheight}{-12cm}   





%
%
%
%
%
%
\bibliographystyle{IEEEtran}
\bibliography{abhi_cdc}

\begin{thebibliography}{10}
\providecommand{\url}[1]{#1}
\csname url@samestyle\endcsname
\providecommand{\newblock}{\relax}
\providecommand{\bibinfo}[2]{#2}
\providecommand{\BIBentrySTDinterwordspacing}{\spaceskip=0pt\relax}
\providecommand{\BIBentryALTinterwordstretchfactor}{4}
\providecommand{\BIBentryALTinterwordspacing}{\spaceskip=\fontdimen2\font plus
\BIBentryALTinterwordstretchfactor\fontdimen3\font minus \fontdimen4\font\relax}
\providecommand{\BIBforeignlanguage}[2]{{%
\expandafter\ifx\csname l@#1\endcsname\relax
\typeout{** WARNING: IEEEtran.bst: No hyphenation pattern has been}%
\typeout{** loaded for the language `#1'. Using the pattern for}%
\typeout{** the default language instead.}%
\else
\language=\csname l@#1\endcsname
\fi
#2}}
\providecommand{\BIBdecl}{\relax}
\BIBdecl

\bibitem{mnih2013playing}
V.~Mnih, K.~Kavukcuoglu, D.~Silver, A.~Graves, I.~Antonoglou, D.~Wierstra, and M.~Riedmiller, ``Playing atari with deep reinforcement learning,'' \emph{arXiv preprint arXiv:1312.5602}, 2013.

\bibitem{kober2013reinforcement}
J.~Kober, J.~A. Bagnell, and J.~Peters, ``Reinforcement learning in robotics: A survey,'' \emph{The International Journal of Robotics Research}, vol.~32, no.~11, pp. 1238--1274, 2013.

\bibitem{wisniewski2017stochastic}
R.~Wisniewski and M.~L. Bujorianu, ``Stochastic safety analysis of stochastic hybrid systems,'' in \emph{56th Annual Conference on Decision and Control (CDC)}.\hskip 1em plus 0.5em minus 0.4em\relax IEEE, 2017, pp. 2390--2395.

\bibitem{wisniewski2020p}
R.~Wisniewski, M.~L. Bujorianu, and C.~Sloth, ``$ p $-safe analysis of stochastic hybrid processes,'' \emph{IEEE Transactions on Automatic Control}, vol.~65, no.~12, pp. 5220--5235, 2020.

\bibitem{bujorianu2020stochastic}
M.~L. Bujorianu, R.~Wisniewski, and E.~Boulougouris, ``Stochastic safety for {M}arkov chains,'' \emph{IEEE Control Systems Letters}, vol.~5, no.~2, pp. 427--432, 2020.

\bibitem{wisniewski2021safety}
R.~Wisniewski and M.~L. Bujorianu, ``Safety of stochastic systems: An analytic and computational approach,'' \emph{Automatica}, vol. 133, p. 109839, 2021.

\bibitem{wisniewski2023probabilistic}
------, ``Probabilistic safety guarantees for {M}arkov decision processes,'' \emph{IEEE Transactions on Automatic Control}, 2023.

\bibitem{mazumdar2023online_ecc}
A.~Mazumdar, R.~Wisniewski, and M.~L. Bujorianu, ``Online learning of safety function for {M}arkov decision processes,'' in \emph{European Control Conference (ECC)}.\hskip 1em plus 0.5em minus 0.4em\relax IEEE, 2023, pp. 1--6.

\bibitem{mazumdar2023online}
------, ``Online model-free safety verification for {M}arkov decision processes without safety violation,'' \emph{to be presented in European Control Conference (ECC)}, 2024.

\bibitem{el2018controlled}
M.~El~Chamie, Y.~Yu, B.~A{\c{c}}{\i}kme{\c{s}}e, and M.~Ono, ``Controlled {M}arkov processes with safety state constraints,'' \emph{IEEE Transactions on Automatic Control}, vol.~64, no.~3, pp. 1003--1018, 2018.

\bibitem{garcia2015comprehensive}
J.~Garc{\i}a and F.~Fern{\'a}ndez, ``A comprehensive survey on safe reinforcement learning,'' \emph{Journal of Machine Learning Research}, vol.~16, no.~1, pp. 1437--1480, 2015.

\bibitem{jansen2020safe}
N.~Jansen, B.~K{\"o}nighofer, S.~Junges, A.~Serban, and R.~Bloem, ``Safe reinforcement learning using probabilistic shields,'' in \emph{31st International Conference on Concurrency Theory}.\hskip 1em plus 0.5em minus 0.4em\relax Schloss Dagstuhl-Leibniz-Zentrum fur Informatik GmbH, Dagstuhl Publishing, 2020, pp. 31--316.

\bibitem{bastani2021safe}
O.~Bastani, ``Safe reinforcement learning with nonlinear dynamics via model predictive shielding,'' in \emph{American control conference (ACC)}.\hskip 1em plus 0.5em minus 0.4em\relax IEEE, 2021, pp. 3488--3494.

\bibitem{altman1991adaptive}
E.~Altman and A.~Schwartz, ``Adaptive control of constrained {M}arkov chains,'' \emph{IEEE Transactions on Automatic Control}, vol.~36, no.~4, pp. 454--462, 1991.

\bibitem{jaksch2010near}
T.~Jaksch, R.~Ortner, and P.~Auer, ``Near-optimal regret bounds for reinforcement learning,'' \emph{Journal of Machine Learning Research}, vol.~11, pp. 1563--1600, 2010.

\bibitem{efroni2020exploration}
Y.~Efroni, S.~Mannor, and M.~Pirotta, ``Exploration-exploitation in constrained mdps,'' \emph{arXiv preprint arXiv:2003.02189}, 2020.

\bibitem{singh2022learning}
R.~Singh, A.~Gupta, and N.~Shroff, ``Learning in constrained {M}arkov decision processes,'' \emph{IEEE Transactions on Control of Network Systems}, 2022.

\bibitem{bura2022dope}
A.~Bura, A.~HasanzadeZonuzy, D.~Kalathil, S.~Shakkottai, and J.-F. Chamberland, ``{DOPE}: Doubly optimistic and pessimistic exploration for safe reinforcement learning,'' \emph{Advances in Neural Information Processing Systems}, vol.~35, pp. 1047--1059, 2022.

\bibitem{wachi2020safe}
A.~Wachi and Y.~Sui, ``Safe reinforcement learning in constrained {M}arkov decision processes,'' in \emph{International Conference on Machine Learning}.\hskip 1em plus 0.5em minus 0.4em\relax PMLR, 2020, pp. 9797--9806.

\bibitem{wei2022triple}
H.~Wei, X.~Liu, and L.~Ying, ``Triple-q: A model-free algorithm for constrained reinforcement learning with sublinear regret and zero constraint violation,'' in \emph{International Conference on Artificial Intelligence and Statistics}.\hskip 1em plus 0.5em minus 0.4em\relax PMLR, 2022, pp. 3274--3307.

\bibitem{wei2022provably}
------, ``A provably-efficient model-free algorithm for infinite-horizon average-reward constrained {M}arkov decision processes,'' in \emph{Proceedings of the AAAI Conference on Artificial Intelligence}, vol.~36, no.~4, 2022, pp. 3868--3876.

\bibitem{wei2023provably}
H.~Wei, A.~Ghosh, N.~Shroff, L.~Ying, and X.~Zhou, ``Provably efficient model-free algorithms for non-stationary cmdps,'' in \emph{International Conference on Artificial Intelligence and Statistics}.\hskip 1em plus 0.5em minus 0.4em\relax PMLR, 2023, pp. 6527--6570.

\bibitem{yu2019convergent}
M.~Yu, Z.~Yang, M.~Kolar, and Z.~Wang, ``Convergent policy optimization for safe reinforcement learning,'' \emph{Advances in Neural Information Processing Systems}, vol.~32, 2019.

\bibitem{ding2020natural}
D.~Ding, K.~Zhang, T.~Basar, and M.~Jovanovic, ``Natural policy gradient primal-dual method for constrained {M}arkov decision processes,'' \emph{Advances in Neural Information Processing Systems}, vol.~33, pp. 8378--8390, 2020.

\bibitem{papini2022smoothing}
M.~Papini, M.~Pirotta, and M.~Restelli, ``Smoothing policies and safe policy gradients,'' \emph{Machine Learning}, vol. 111, no.~11, pp. 4081--4137, 2022.

\bibitem{guin2023policy}
S.~Guin and S.~Bhatnagar, ``A policy gradient approach for finite horizon constrained {M}arkov decision processes,'' in \emph{62nd Annual Conference on Decision and Control (CDC)}.\hskip 1em plus 0.5em minus 0.4em\relax IEEE, 2023, pp. 3353--3359.

\bibitem{haviv1996constrained}
M.~Haviv, ``On constrained {M}arkov decision processes,'' \emph{Operations research letters}, vol.~19, no.~1, pp. 25--28, 1996.

\bibitem{maurer2009empirical}
A.~Maurer and M.~Pontil, ``Empirical bernstein bounds and sample variance penalization,'' \emph{arXiv preprint arXiv:0907.3740}, 2009.

\end{thebibliography}
\end{document}